\documentclass[10pt, a4paper]{article}

\usepackage[final]{lrec2026} 

\usepackage{graphicx}
\usepackage{multirow}
\usepackage{tabularx}
\usepackage{booktabs}
\usepackage{amsmath}
\usepackage[table]{xcolor}
\usepackage{tcolorbox}
\usepackage{makecell}
\usepackage{amsmath}
\usepackage{amssymb}
\DeclareMathOperator{\argTopK}{arg-TopK}

\title{Merging Continual Pretraining Models for Domain-Specialized LLMs: A Case Study in Finance}

\name{$^\dagger$Kentaro Ueda, $^\ddagger$François Portet, $^\dagger$Hirohiko Suwa, $^\dagger$Keiichi Yasumoto} 

\address{$^\dagger$NARA Institute of Science and Technology, $^\ddagger$Université Grenoble Alpes \\
         \{ueda.kentaro.ug2, h-suwa, yasumoto\}@is.naist.jp\\
         francois.portet@imag.fr\\
         }

\newcommand{\kentaro}[1]{\textcolor{black}{#1}}

\abstract{
While LLMs excel at general tasks, they struggle in specialized domains like finance, requiring diverse skills in domain knowledge, mathematical reasoning, and multilingual processing. Merging domain-specific Continual Pre-training (CPT) "experts" offers a practical alternative to costly and unstable multi-skill training. However, unlike established Supervised Fine-Tuning (SFT) model-based merging, CPT model merging remains largely unexplored. We address this gap by creating financial LLMs from experts in finance, math, and Japanese.
We propose a three-stage evaluation focusing on knowledge recovery, complementarity, and emergence, and assess three merging methods (Task Arithmetic, TIES, and DARE-TIES) on a comprehensive financial benchmark curated from 18 tasks across 8 established datasets.
Results show that merging an expert with its base model recovers general knowledge lost during CPT, while merging experts improves performance and can yield emergent cross-domain skills. Among the methods, Task Arithmetic performs strongly but is hyperparameter-sensitive, whereas TIES is more robust. Our findings also suggest that while model similarity correlates with merging success, emergent skills depend on more complex factors. This work presents the first foundational analysis of CPT model merging, establishing a principled framework and providing clear guidance for building multi-skill LLMs from existing assets.
 \\ \newline }

\begin{document}
\maketitleabstract

\section{Introduction}
\label{sec:introduction}
Large Language Models (LLMs) demonstrate strong general-purpose NLP capabilities, but often underperform in specialized domains such as finance\cite{finance_surveys, srivastava-etal-2024-evaluating, Wu2023BloombergGPTAL}, where a combination of heterogeneous skills is required—including domain-specific terminology, mathematical reasoning, tabular comprehension, and multilingual processing. These skills are rarely acquired through standard pretraining, limiting LLM performance on complex financial tasks. This issue persists to some extent even with API-accessible LLMs trained on highly diverse data, and in practice, their use is further constrained by \kentaro{security and privacy requirements.}

Continual pretraining (CPT) and supervised fine-tuning (SFT) have been the dominant approaches to address this problem\cite{cheng-etal-2024-instruction, yang2023fingpt}. While CPT enriches models with domain-wide knowledge via large-scale corpora, and SFT specializes models on task-specific data, both approaches have limitations. Directly creating a comprehensive financial expert model that addresses these needs requires training on diverse datasets across multiple domains under substantial computational resources, all while risking catastrophic forgetting\cite{thompson-etal-2019-overcoming} and overfitting. 

An alternative paradigm is model merging\citep{sung-etal-2023-empirical, yang2024model_survey}, which composes a financial expert from existing, pre-trained models. This technique enables the fusion of specialized skills from each constituent model without requiring costly and unstable retraining. This approach not only saves significant computational resources but also represents a powerful form of reusing existing language resources, specifically pre-trained models. This means that it assembles new, high-value capabilities from existing assets.

\begin{figure}[t]
  \centering
  \includegraphics[width=\columnwidth]{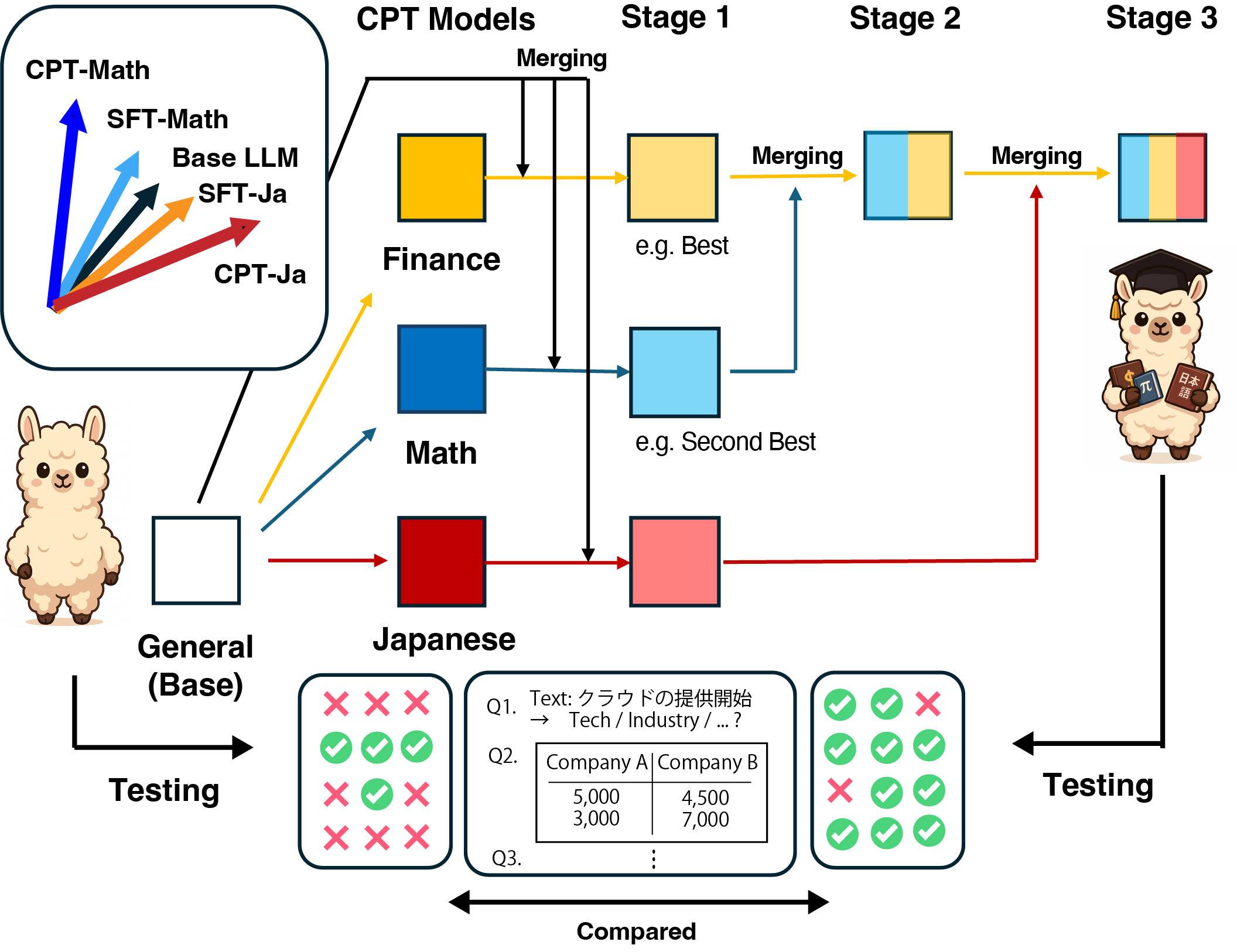}
  \caption{Three-stage model merging framework designed to analyze knowledge transfer, complementarity, and interference in CPT model integration.}
  \label{fig:overview}
\end{figure}

\kentaro{While this approach is promising, most existing work has focused on merging SFT models. To the best of our knowledge, there is no empirical research on CPT model merging.}
Because CPT models are imparted with broad domain-level knowledge, they diverge more significantly in parameter space than SFT models, which are injected with localized task knowledge\footnote{In our preliminary experiments, we also confirmed this divergence via L2 distance and cosine similarity. Further details \kentaro{will be included} in the Appendix \ref{sec:distance}}. Therefore, merging CPT models can lead to destructive interference during integration, posing a new question of how their knowledge interacts in weight space.

Some recent studies have taken steps toward merging CPT models, such as Branch-Train-MiX~\cite{sukhbaatar2024branch} and Chat Vector~\cite{huang-etal-2024-chat}. However, these rely on specialized architectures or constrained settings and do not provide a systematic evaluation of CPT-model merging in LLMs.

In this work, we conduct the first foundational study of CPT-based model merging, introducing a principled, three-stage framework designed to assess knowledge recovery, complementarity, and emergence. Within this framework, we augment a powerful open-source LLM as a base by fusing it with three publicly available, open-weight CPT models specialized in finance knowledge, mathematical reasoning, and Japanese language processing, combining their unique expertise. 

\kentaro{We compare three representative merging methods, Task Arithmetic (TA)\cite{task_arithmetic}, TIES (TI)\cite{ties}, and DARE-TIES (DA)\cite{dare}, by running the three-stage pipeline once per method, keeping the method fixed across stages, and sweeping hyperparameters at each stage.}

We further design a comprehensive financial benchmark by curating 18 tasks from 8 public datasets to cover the diverse skills required.

Our results show that base+CPT merging consistently outperforms the constituent CPT models, recovering general knowledge lost during CPT. Furthermore, merging between CPT models, such as Finance+Math, can yield emergent capabilities on complex reasoning tasks. Notably, the inclusion of the Japanese CPT model even boosts performance on English tasks, suggesting cross-lingual spillover effects. Among the merging methods, TA achieves the highest gains but is sensitive to tuning, while TI offers more robust performance. 

\kentaro{Higher similarity between the constituent models (e.g., lower L2 distance or higher cosine similarity) generally leads to larger gains from merging. However, whether the merged model can surpass the best constituent model’s performance cannot be inferred from similarity alone.}

Our analysis reveals that CPT model merging is not only viable but is an effective and resource-efficient paradigm that can produce models exceeding the performance of their constituents, offering a new path for building multi-skilled LLMs.

\section{Related work}
\label{sec:related_work}
\subsection{Domain Adaptation for Financial LLMs}
The primary methods for adapting LLMs to the financial domain are Supervised Fine-tuning (SFT) and Continual Pre-training (CPT)\cite{xie2023pixiu, yang2023fingpt}. While CPT is powerful for instilling domain knowledge, it demands prohibitive computational resources. Furthermore, both approaches risk ``catastrophic forgetting \cite{thompson-etal-2019-overcoming},'' which is the tendency to overfit to new data at the expense of general capabilities. Consequently, integrating multiple distinct skills into a single model via these methods poses significant challenges in both cost and stability.

\subsection{Model Merging Techniques}
Model merging, a technique for combining the knowledge and capabilities of multiple models without additional training, has recently garnered attention \cite{yang2024model_survey, sung-etal-2023-empirical, yadav2025what}. Prominent methods such as Task Arithmetic \cite{task_arithmetic}, TIES \cite{ties}, and DARE \cite{dare} were designed primarily for merging SFT-created experts and consequently assume small parameter divergence. This makes them potentially unsuitable for CPT models, which tend to exhibit large weight shifts from their base model.

Several studies have begun exploring the CPT model merging. For example, Branch-Train-MiX \citep{sukhbaatar2024branch} introduces a mixture-of-experts-style pre-branching mechanism; Chat Vector \citep{huang-etal-2024-chat} investigates language adaptation via merging; and other works apply merging to mitigate catastrophic forgetting \citep{siriwardhana2024domain}. Additionally, prior efforts have explored instruction-following fine-tuning on top of financial CPT models \citep{hirano2024construction}. However, all of these approaches rely on task-specific or architecture-specific designs, and, to the best of our knowledge, a general and systematic investigation into CPT model merging has not yet been conducted. 

\subsection{Evaluation Benchmarks for Financial NLP}
Numerous benchmarks have been proposed to measure the capabilities of financial LLMs, targeting skills such as sentiment analysis (FPB \cite{FPB_paper}), quantitative QA (ConvFinQA \cite{chen-etal-2022-convfinqa}), and document parsing (DocMath \cite{zhao-etal-2024-docmath}). While valuable for assessing these isolated skills, their structure is ill-suited for analyzing the complex effects of model merging. The core challenge in evaluating merging is to create a detailed ``skill profile''—tracking which capabilities are enhanced, which are degraded, and which remain stable. Existing frameworks, which often reduce performance to a single aggregate score, lack the necessary granularity and breadth for such a multifaceted analysis. Therefore, a comprehensive benchmark designed to provide a fine-grained investigation of these skill-level interactions has been a critical missing resource. This gap motivated our work to curate 18 tasks from 8 public datasets.

\section{Three-stage Framework for Analyzing CPT Model Merging}
Our analysis of CPT-model merging is guided by a principled, three-stage framework (Figure~\ref{fig:overview}). This hierarchical methodology is designed to make the analysis tractable, circumventing the combinatorial explosion of an exhaustive search. \kentaro{The output model from each stage is carried over to form the basis of the next stage.} Across these stages, we comprehensively evaluate the key effects of merging, including knowledge recovery, cross-domain complementarity, and emergent behavior.

\subsection{Model Setup and Constituent CPT Models}
We use \texttt{Llama-3-8B}\footnote{\texttt{meta-llama/Meta-Llama-3-8B}} \cite{llama3modelcard} as the base model $\theta_0$, and three CPT models targeting key skills in financial NLP: CPT-Finance\footnote{\texttt{instruction-pretrain/finance-Llama3-8B}}\cite{cheng-etal-2024-instruction} for financial domain knowledge, CPT-Math\footnote{\texttt{MathGenie/MathCoder2-Llama-3-8B}}\cite{lu2024mathcoder2bettermathreasoning} for mathematical reasoning, and CPT-Japanese\footnote{\texttt{tokyotech-llm/Llama-3-Swallow-8B-v0.1}}\cite{Fujii:COLM2024} for multilingual (Japanese) understanding. All three models were trained via CPT by independent groups from the same base checkpoint $\theta_0$ and publicly released; we did not perform any additional pretraining and only used the open-weight checkpoints downloaded from Hugging Face. These models differ in training data, training objectives, and language coverage, representing distinct knowledge axes and inducing varying parameter shifts from $\theta_0$. 

\subsection{Stage 1: Single-CPT Integration for Isolated Transfer Analysis}
In Stage 1, each CPT model ($\theta_f$ (Finance), $\theta_m$ (Math), and $\theta_j$ (Japanese)) is individually merged with the base model $\theta_0$.
This design eliminates cross-CPT interference and measures each CPT’s merging effect (i.e., performance impact). In addition, by quantifying the degree of knowledge transfer from each CPT to the base, we assess how much general capability lost during continual pretraining (catastrophic forgetting) is recovered.

\kentaro{For each merge method $a$, hyperparameter $\gamma$ (e.g. merging weight) and CPT domain $c$,} the resulting Stage 1 model, denoted as $\theta^{(1)}_{a,c}(\gamma)$, is generated by:
\begin{equation}
    \theta^{(1)}_{a,c}(\gamma) = f_a(\theta_0, \theta_c; \gamma)
\end{equation}
where $c \in \{f,m,j\}$, $\theta_c$ represents the CPT model weights, and $\gamma \in \Gamma_a$ is \kentaro{a hyperparameter of the merging method $a$.}

\subsection{Stage 2: Dual-CPT Integration for Complementarity Analysis}
In Stage 2, we construct a dual-CPT model by merging the two strongest single-CPT models from Stage 1, originating from distinct CPT domains, to study cross-domain complementarity and emergent behavior.

First, for each merge method $a$ and CPT domain $c \in \{f,m,j\}$, we identify the best-performing Stage 1 model, denoted as $\theta^{*(1)}_{a,c}$. This is achieved by selecting the optimal hyperparameter $\hat{\gamma}_{a,c}$ that maximizes the score $\mathcal{S}$ \kentaro{(e.g., a benchmark score)}:
\begin{equation}
\begin{split}
\theta^{*(1)}_{a,c} = \theta^{(1)}_{a,c}(\hat{\gamma}_{a,c}),\\
\text{where} \quad
\hat{\gamma}_{a,c} = \operatorname*{arg\,max}_{\gamma \in \Gamma_a} \mathcal{S}\!\left(\theta^{(1)}_{a,c}(\gamma)\right)
\end{split}
\end{equation}
Next, we select the top two domains, denoted by the set $C^*_a$, based on their scores on:
\begin{equation}
    C^*_m = \argTopK_{c \in \{f,m,j\}} \left( \mathcal{S}(\theta^{*(1)}_{a,c}), K=2 \right)
\end{equation}
Finally, we merge the two selected models to form the Stage 2 model, $\theta^{(2)}_a$, using a new hyperparameter $\gamma' \in \Gamma_a$:

\begin{equation}
\begin{split}
    \theta^{(2)}_a(\gamma') = f_a(\theta^{*(1)}_{a,c_1^*}, \theta^{*(1)}_{a,c_2^*}; \gamma'),\\
    \text{where} \quad \{c_1^*, c_2^*\} = C^*_a
\end{split}
\end{equation}
This dual integration reveals whether domain-specific capabilities (e.g., mathematical reasoning and financial knowledge) enhance each other synergistically or interfere destructively.

\subsection{Stage 3: Full-CPT Integration for Capacity Scaling Evaluation}
In Stage 3, we build a full-CPT model that integrates all three domains to evaluate whether the additional integration yields further gains or leads to degradation due to increased interference. This stage analyzes the scalability and integration limits of the merging methods.

First, we identify the best-performing Stage 2 model, $\theta^{*(2)}_a$, by optimizing its hyperparameter $\gamma'$:
\begin{equation}
\begin{split}
    \theta^{*(2)}_a = \theta^{(2)}_a(\hat{\gamma}'_a),\\ \quad \text{where} \quad \hat{\gamma}'_a = \operatorname*{arg\,max}_{\gamma' \in \Gamma_a} \mathcal{S}\left( \theta^{(2)}_a(\gamma') \right)
\end{split}
\end{equation}
Let $c_3^*$ be the remaining domain not selected in Stage 2, i.e., $\{c_3^*\} = \{f,m,j\} \setminus C^*_a$. The full-CPT model, $\theta^{(3)}_a$, is then constructed by merging the best Stage 2 model with the remaining best Stage 1 model:
\begin{equation}
    \theta^{(3)}_a(\gamma'') = f_a(\theta^{*(2)}_a, \theta^{*(1)}_{a,c_3^*}; \gamma'')
\end{equation}
where $\gamma'' \in \Gamma_a$ is a new hyperparameter for the final merge.

\subsection{Merge Techniques for CPT Integration}
To comprehensively evaluate the effectiveness of CPT model integration, we adopt three representative merge methods: Task Arithmetic (TA)\cite{task_arithmetic}, TIES-Merging (TI)\cite{ties}, and DARE-TIES (DA)\cite{dare}. These methods span a spectrum from simple linear combination to sign-based selective integration and stochastic interference mitigation, thus providing a broad perspective on the merge design space.
\textbf{Task Arithmetic (TA)} is a linear merge strategy that adds scaled task vectors to the base model, where a task vector represents the weight difference between a CPT model and the base model.
Given a pretrained base model $\theta_0$ and a CPT model $\theta_t$ fine-tuned on domain $t$, the task vector is defined as $\tau_t = \theta_t - \theta_0$, and the merged model is computed as $\theta_{\text{final}} = \theta_0 + \sum_{t=1}^{T} \lambda_t \cdot \tau_t$.

Here, $\lambda_t$ is the scaling coefficient for task vector $\tau_t$. TA assumes linear compositionality in weight space, offering a simple yet surprisingly effective approach for knowledge integration.

\textbf{TIES-Merging (TI)} performs sparse and sign-consistent integration. For each task vector $\tau_t$, only the top-$d$ \% of elements (by magnitude) are retained, and the rest are zeroed out to obtain a pruned vector $\hat{\tau}_t$. Then, for each parameter $p$, we compute the representative sign $\gamma_{\text{maj}}^p$ by taking a majority vote over the normalized deltas:
$\gamma_{\text{maj}}^p = \text{sgn} \left( \sum_{t=1}^{\kentaro{T}} \hat{\tau}_t^p \right)$

The final merged parameter $\theta_{\text{merge}}^p$ is computed by adding to the base weight $\theta_{\text{base}}^p$ the average of normalized deltas $\hat{\tau}_t^p$ from models whose signs match the majority, i.e., $\theta_{\text{merge}}^p = \theta_{\text{base}}^p + \lambda \cdot \frac{1}{|\mathcal{A}^p|} \sum_{t \in \mathcal{A}^p} \hat{\tau}_t^p$, where $\mathcal{A}^p = \left\{ t \in [\kentaro{1,T}] \mid \kentaro{\text{sgn}(\hat{\tau}_{t}^p)} = \gamma_{\text{maj}}^p \right\}$ is the index set of sign-aligned models.

By explicitly excluding sign conflicts, TIES-Merging promotes stable integration by combining only mutually consistent knowledge across CPTs.

\textbf{DARE-TIES (DA)} extends TIES-Merging by incorporating a stochastic Drop-And-ReScale (DARE) preprocessing step~\cite{dare}. For each task vector $\tau_t$, a Bernoulli mask $\mathbf{M}_t \sim \mathrm{Bernoulli}(d)$ with retention rate $d \in (0, 1]$ is applied, and the resulting masked vector is rescaled as $\hat{\tau}_t = \frac{\mathbf{M}_t \odot \tau_t}{d}$ to preserve the expected magnitude. TIES-Merging is then applied to the set $\{\hat{\tau}_t\}$. 
\kentaro{DA achieves improved robustness over TIES alone by dispersing task vector interference while preserving the expected signal magnitude.}

\paragraph{Hyperparameter Sweep}
For each merging method $m$, we sweep its key hyperparameter over a predefined range. 
This corresponds to the general hyperparameter $\gamma$ used in our formulation. 
Specifically, $\gamma$ represents the scaling coefficient $\lambda$ for TA, and the density parameter $d$ for TI and DA. 
For all methods, the hyperparameter candidate set $\Gamma_a$ is defined as the 9 discrete values $\{0.1, 0.2, \dots, 0.9\}$. 
This sweep enables a quantitative analysis of each hyperparameter's impact on merging performance.

\subsection{Benchmark for Evaluating Knowledge Integration in Financial Tasks}
To evaluate CPT-based model merging in a realistic and multi-faceted setting, we construct a benchmark covering 18 tasks from 8 established financial NLP datasets. The benchmark targets four key competencies: (i) sentiment and classification, (ii) information extraction, (iii) multilingual comprehension, and (iv) mathematical reasoning.

\textbf{Sentiment and classification} tasks include FiQA SA~\citep{FiQA_SA}, FPB~\citep{FPB_paper}, and Headline~\citep{Headline_sinha2021impact}, which assess the ability to interpret market sentiment and domain-specific expressions. \textbf{Information extraction} is evaluated with NER~\citep{NER_salinas-alvarado-etal-2015-domain}, focusing on financial entities such as organizations, currencies, and instruments. \textbf{Multilingual comprehension} is assessed using three subsets of Multifin~\citep{jorgensen-etal-2023-multifin}: English-only (En), Japanese-only (Ja), and the full 15-language set (All), enabling fine-grained analysis of cross-lingual generalization. \textbf{Mathematical reasoning} is evaluated via ConvFinQA~\citep{chen-etal-2022-convfinqa} (multi-turn QA over tables), DocMathEval~\citep{zhao-etal-2024-docmath} (with SS/SL/CS complexities), and FinanceMath~\citep{zhao-etal-2024-knowledgefmath}, which covers both tabular and non-tabular formats.
Dataset details \kentaro{will be included} in the Appendix~\ref{sec:dataset}.

\subsection{Evaluation Metrics for Merge Performance}
To measure the primary evaluation axes of this work (knowledge recovery, complementarity, and emergence), we introduce six metrics that provide operational definitions for these phenomena, quantifying synergistic gains and destructive interference to enable a multifaceted analysis of the merging effects.
Let $s_i^{\text{merge}}$ be the score of the merged model on task $i$, $s_i^{(j)}$ the score of constituent model $j$, $k$ the number of constituent models, and $N$ the total number of tasks.

\textbf{Gain:}
For each task $i$, we define Gain as the difference between the merged model's score and the average score of the constituent models:  
$\text{Gain}_i = s_i^{\text{merge}} - \frac{1}{k} \sum_{j=1}^{k} s_i^{(j)}$.

\textbf{Macro-Gain:}
Macro-Gain is the average of Gain across all tasks, computed as  
$\text{Macro-Gain} = \frac{1}{N} \sum_{i=1}^{N} \text{Gain}_i$.  
This metric captures the overall performance improvement of the merged model relative to its components.

\textbf{Outperform Gap (OG):}
OG quantifies whether the merged model outperforms the best of its constituent models for task $i$, defined as  
$\text{OG}_i = s_i^{\text{merge}} - \max_{1 \leq j \leq k} s_i^{(j)}$.  
A positive OG indicates knowledge complementarity, while a negative OG suggests interference.

\textbf{Macro-OG:}
Macro-OG is the average OG across all tasks:  
$\text{Macro-OG} = \frac{1}{N} \sum_{i=1}^{N} \text{OG}_i$.  
It provides a summary of the overall tendency for knowledge fusion or conflict.

\textbf{Oracle Retention:}
This metric measures how closely the merged model's aggregate performance approaches a theoretical upper bound. This upper bound is defined as the "Oracle Score"—the total score achieved by an ideal oracle that selects the best-performing constituent model for each individual task. It is calculated as: 
$\text{OracleRetention} = \frac{\sum_{i=1}^{N} s_i^{\text{merge}}}{\sum_{i=1}^{N} \max_{1 \leq j \leq k} s_i^{(j)}}$.  
A value of 1.0 indicates that the merged model has perfectly replicated the oracle's performance in aggregate, successfully retaining the peak capabilities of all its constituents. 
A value greater than 1.0 signifies that the merged model's total score has surpassed this strong theoretical upper bound.

\kentaro{\textbf{Overall:}}
\kentaro{The overall score represents the macro-average performance across all benchmark tasks. }

\section{Experimental Results and Analysis}
\subsection{Stage-wise Performance Impact of Model Merging}
Table~\ref{tab:three_stage_results} presents the quantitative results of CPT model integration for building financial LLMs. 

\begin{table*}[t]
\centering
\scriptsize
\label{tab:three_stage_results}
\begin{tabular}{ccccccc}
\toprule
\multirow{1}{*}{\textbf{Model ID}} & 
\multirow{1}{*}{\textbf{Model}} & 
\multirow{1}{*}{\textbf{Method}} & 
\multirow{1}{*}{\textbf{Overall}} & 
\multirow{1}{*}{\textbf{Macro-Gain}} & 
\multirow{1}{*}{\textbf{Macro-OG}}  &
\multirow{1}{*}{\textbf{OracleRetention}} \\

\midrule
B & Base   &  -   &  24.88 & –  & -  & - \\
F & CPT-Finance &  - & 35.89 & –  & -  & -\\
M & CPT-Math    &   -  & 23.85 & –  & - & -\\
J & CPT-JP      &   -  & 25.29 & –  & - & - \\
\midrule
\rowcolor{gray!10}
\multicolumn{7}{c}{\textbf{Stage 1}} \\
\text{BF$_{\text{TA}}$} & B + F        & TA $(\lambda_{\text{best}}=0.7)$ & 36.50 &  6.12 & 0.11 & 1.00 \\
\text{BM$_{\text{TA}}$} & B + M        & TA $(\lambda_{\text{best}}=0.6)$  & 27.43 & 2.87 & 1.80 &1.07\\
\text{BJ$_{\text{TA}}$} & B + J        & TA $(\lambda_{\text{best}}=0.8)$& 26.00 & 1.11 & -0.08 & 1.00 \\
\midrule
\text{BF$_{\text{TI}}$} & B + F        & TI $(d_{\text{best}}=0.3)$& 35.72 & 5.34 &  -0.67  &0.98  \\
\text{BM$_{\text{TI}}$} & B + M        & TI $(d_{\text{best}}=0.3)$& 26.49 & 1.93 &  0.86 & 1.03\\
\text{BJ$_{\text{TI}}$} & B + J        & TI $(d_{\text{best}}=0.4)$& 26.16 & 1.27 &   0.07 & 1.00\\
\midrule
\text{BF$_{\text{DA}}$} & B + F        & DA $(d_{\text{best}}=0.8)$ & 34.49 & 4.10 &  -1.91 & 0.95  \\
\text{BM$_{\text{DA}}$} & B + M        & DA $(d_{\text{best}}=0.9)$& 25.15 & 0.59 &  -0.48 & 0.98 \\
\text{BJ$_{\text{DA}}$} & B + J        & DA $(d_{\text{best}}=0.9)$& 25.61 & 0.72 &  -0.47 & 0.98 \\
\midrule
\rowcolor{gray!10}
\multicolumn{7}{c}{\textbf{Stage 2}} \\
\text{FM$_{\text{TA}}$} & \text{BF$_{\text{TA}}$} + \text{BM$_{\text{TA}}$}    &  TA $(\lambda_{\text{best}}=0.8)$ & 38.92 & 6.96 & 2.02 & 1.05 \\
\text{FM$_{\text{TI}}$} & \text{BF$_{\text{TI}}$} + \text{BM$_{\text{TI}}$}    &  TI $(d_{\text{best}}=0.2)$ & 38.39 & 7.29 & 2.11 & 1.06\\
\text{FJ$_{\text{DA}}$} & \text{BF$_{\text{DA}}$} + \text{BJ$_{\text{DA}}$}    &  DA $(d_{\text{best}}=0.9)$ & 33.78 & 3.73 & -1.17 & 0.97 \\
\midrule
\rowcolor{gray!10}
\multicolumn{7}{c}{\textbf{Stage 3}} \\
\text{FMJ$_{\text{TA}}$} & \text{FM$_{\text{TA}}$} + \text{BJ$_{\text{TA}}$}    &  TA $(\lambda_{\text{best}}=0.8)$ & 37.82 & 5.36 & -1.25 & 0.97 \\
\text{FMJ$_{\text{TI}}$} & \text{FM$_{\text{TI}}$} + \text{BJ$_{\text{TI}}$}    &  TI $(d_{\text{best}}=0.9)$ & 31.56 & -0.72 &  -7.53 & 0.81 \\
\text{FJM$_{\text{DA}}$} & \text{FJ$_{\text{DA}}$} + \text{BM$_{\text{DA}}$}    &  DA $(d_{\text{best}}=0.9)$ & 24.86 & -4.61 & -9.34 & 0.73 \\
\bottomrule
\end{tabular}
\caption{Three-stage evaluation of CPT model merging. Results shown are for the best-performing hyperparameters; \kentaro{A detailed, task-level breakdown of the hyperparameter sweep will be included in the Appendix \ref{sec:skill-level-results}.} The reported values are the averages over three runs with different random seeds. The maximum variance among the results was 0.048, and the average variance computed across models was 0.01, suggesting that the impact of the random seed on the results is limited.}
\end{table*}

\subsubsection{Stage 1: Integrating Base and Single CPT Model}
In Stage 1, each CPT model is individually merged with the base model. As shown in Table~\ref{tab:three_stage_results}, all merged models exhibit positive Macro-Gain, indicating a consistent performance uplift over the constituent average. 
This provides initial evidence that merging can recover general capabilities lost during continual pretraining. 
This conclusion is further strengthened by models like BF\textsubscript{TA}\footnote{This defines the Model IDs shown in the table. For example, BF\textsubscript{TA} indicates a model created by merging B (Base) and F (CPT-Finance) using the TA method} and BM\textsubscript{TA}, which achieves a positive Macro-OG. These findings indicate that merging a strengthened CPT model with its base is worthwhile, since it can both recover forgotten knowledge and unlock further improvements.

\subsubsection{Stage 2: Integrating Domain-Expert CPT Models}
In Stage~2, the integration of the Finance and Math CPT models yielded notable knowledge emergence. Specifically, the merged models FM\textsubscript{TA} and FM\textsubscript{TI} achieved a Macro-OG exceeding 2.
This suggests that the combination of financial knowledge and mathematical reasoning may have interacted complementarily in weight space, facilitating the emergence of new capabilities beyond those present in the individual models.

\subsubsection{Stage 3: Multi-CPT Merging Effects}
In Stage~3, we extend dual-CPT merged models by adding the remaining CPT model, creating tri-domain configurations. Unlike earlier stages, all tri-merged models show a decline in Overall performance, and Oracle Retention falls below 1.0 in most cases.

This suggests that integrating three \kentaro{specialized} CPTs introduces significant interference or redundancy, limiting the scalability of merging strategies. 
These results highlight that while merging two CPTs can yield complementarity and emergent capabilities, adding a third may lead to diminishing returns. This finding emphasizes the need for more selective or modular integration approaches when combining multiple specialized models.

\begin{figure*}[h]
  \centering
  \includegraphics[width=0.9\textwidth]{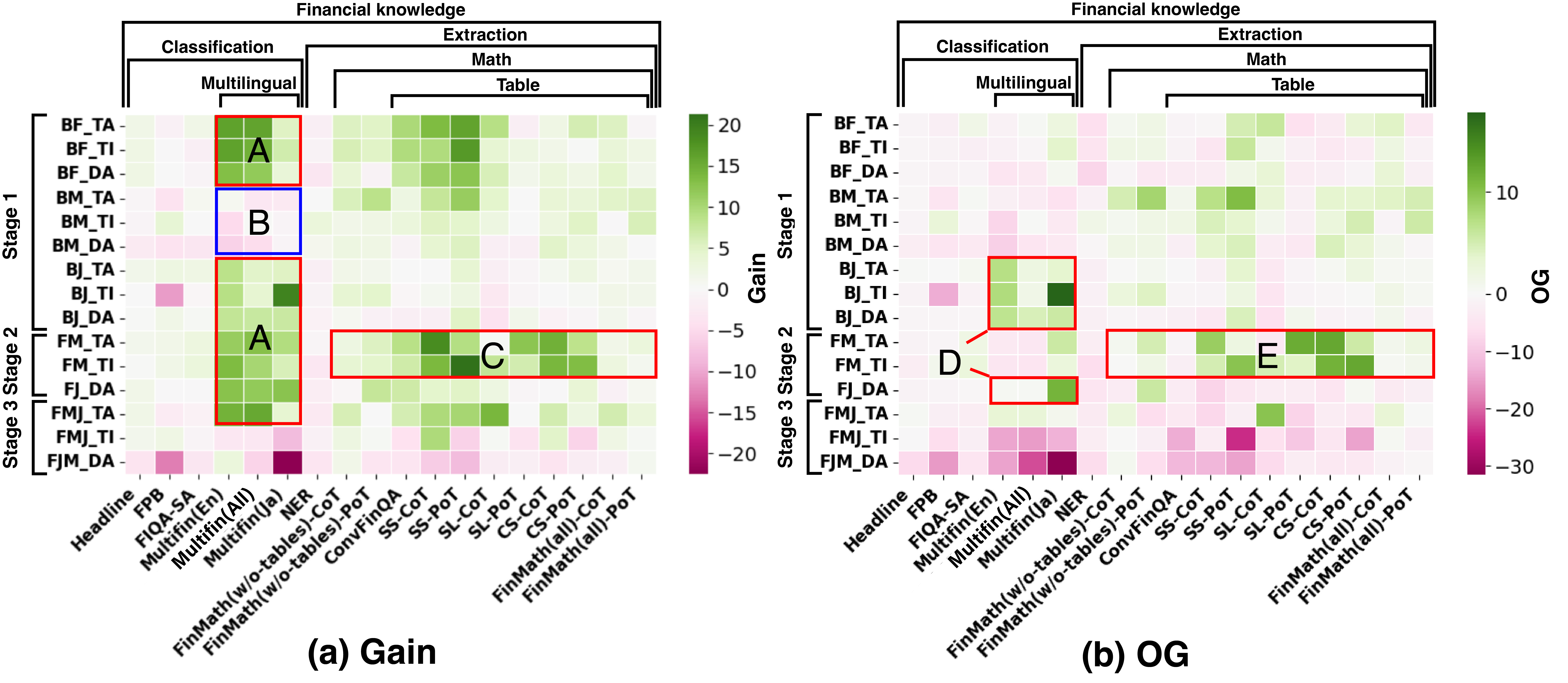}
  \caption{
Task-level effects of CPT-based model merging. The left heatmap shows Gain (improvement over the constituent average), and the right shows Outperform Gap (OG; improvement over the best constituent).}
  \label{fig:og}
\end{figure*}

\subsection{Deconstructing Complementarity and Interference at the Task Level}
Across the three merging stages, we observe diverse knowledge dynamics arising from CPT-based model merging, including \emph{recovery}, \emph{complementarity}, \emph{emergence}, and \emph{interference}. 
Figure~\ref{fig:og} provides a task-level visualization of these dynamics by plotting two key metrics. Gain (left) is the performance improvement over the average of the constituent models, while the OG (right) is the margin against the best of the constituents.

In \textbf{Region~A}, we find that multilingual text classification tasks (Multifin) consistently benefit from the integration of base and CPT models (e.g., BF and BJ in Stage~1), exhibiting positive Gain. This suggests that language-specific knowledge tends to be robustly retained and even reactivated through integration. Moreover, these gains are preserved in Stage~2 mergers, indicating the relative stability of linguistic knowledge even under heterogeneous model integration.

By contrast, \textbf{Region~B} reveals that integrating Math-oriented CPT models into language-focused systems often results in performance degradation on linguistic tasks. This likely reflects representational conflicts between abstract, symbolic structures (e.g., arithmetic logic) and natural language patterns, leading to destructive interference in parameter space.

In \textbf{Regions~C and~E}, merged models such as FM\textsubscript{TA} and FM\textsubscript{TI}, which combine Finance and Math knowledge, achieve substantial Gain and OG on quantitatively demanding tasks like FinanceMath and DocMathEval.
These improvements are particularly prominent on high-complexity subtasks (e.g., SL and CS), suggesting that integration yields not just additive capabilities but emergent reasoning skills beyond those accessible to the original models. This provides empirical evidence for the non-linear compositionality of complementary knowledge in merged models.

\textbf{Region~D} further reveals that the Japanese-specialized model BJ\textsubscript{TA} improves OG not only on Japanese tasks (Multifin(Ja)) but also on English ones (Multifin(En)), implying a form of cross-lingual generalization. This spillover effect suggests that linguistic knowledge may be flexibly reconstructed and propagated through integration in parameter space.

Taken together, our analysis shows that CPT-based model merging induces not merely additive improvements, but diverse effects that depend on the nature, compatibility, and representational alignment of the integrated knowledge. These findings underscore the potential of model merging as a vehicle for achieving knowledge compositionality in parameter space.

\subsection{Emergent and Degenerated Capabilities: A Qualitative Study}
\label{sec:maintext_qualitative}
Model merging occasionally resulted in \emph{emergent capabilities}, allowing the merged model to solve tasks that neither constituent could handle individually. To better understand this phenomenon, we qualitatively analyze both success and failure cases on the CS-CoT task, which showed the largest Outperform Gaps (OG) across all benchmarks.

\begin{figure}[t]
  \centering
  \includegraphics[width=\columnwidth]{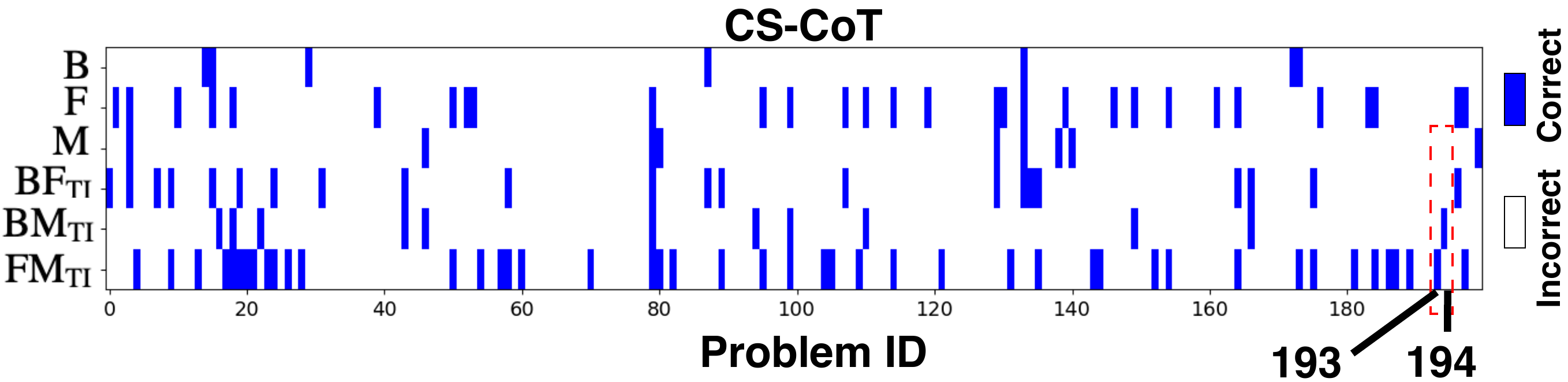}
  \caption{Model-wise answer sheets for the CS-CoT task. \kentaro{(Blue = correct, White = incorrect).}}
  \label{fig:score-sheet}
\end{figure}

Figure~\ref{fig:score-sheet} shows model predictions on CS-CoT, with a focus on two illustrative examples: a successful case (ID193) and a failure case (ID194). These cases highlight both the potential and limitations of CPT-based integration. Full model outputs \kentaro{will be included} in the Appendix~\ref{sec:appendix_outputs}.

\begin{figure}[t]
  \centering
  \includegraphics[width=0.8\columnwidth]{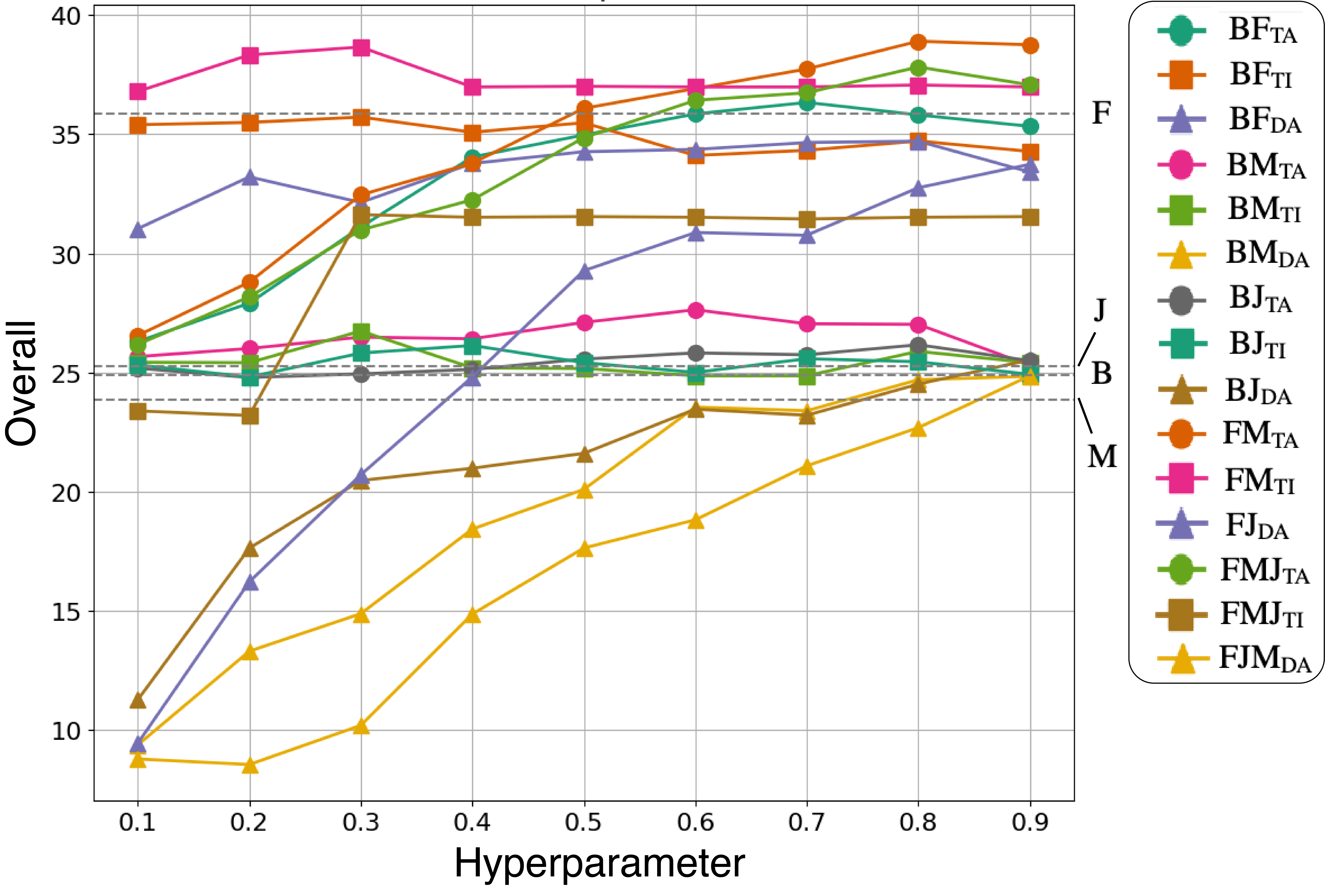}
  \caption{Overall performance across different merge strategies (TA, TI, DA) under varying hyperparameter values. TA sweeps the scaling coefficient $\lambda$, while TI and DA sweep the sparsity/density parameter $d$.}
  \label{fig:overall}
\end{figure}

\begin{figure*}[t]
  \centering
  \includegraphics[width=\textwidth]{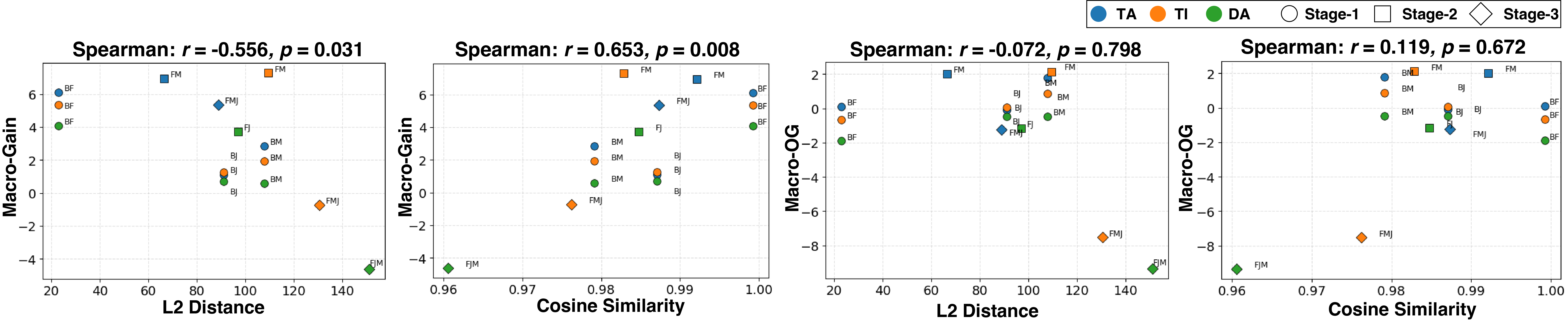}
  \caption{Relationship between the L2 distance and cosine similarity of pre-merge models and the resulting performance (Macro-Gain and Macro-OG). \kentaro{Macro-Gain is higher when the two models are more similar (smaller L2, larger cosine; Spearman’s rank correlation, two-sided $p < 0.05$), but we do not see a clear pattern for Macro-OG ($p > 0.60$).}}
  \label{fig:cosdis}
\end{figure*}

\subsubsection{Successful Case: Emergent Ability via Heterogeneous Knowledge Integration}
For the successful case \kentaro{(ID193)}, the task  involves computing the change in "Non-marketable equity and other investments" using a hypothetical 2018 value (\$20M) and an actual 2017 value (\(-\$46\)M), yielding the correct answer \(20 - (-46) = 66\).

The Finance model (BF) failed by extracting an incorrect 2017 value (\$42) and outputting \$5, suggesting errors in table parsing. The Math model (BM) produced irrelevant reasoning about unrelated entities, failing to engage with the question.

In contrast, the merged model (FM) correctly retrieved both values and executed the equation with clear arithmetic reasoning. This success reflects a complementary integration: BF's document grounding and BM's symbolic reasoning were jointly activated to solve a task that neither model could handle alone.

\kentaro{This case highlights a genuine emerging ability, where isolated financial and mathematical knowledge join together to form a composite reasoning skill.}

\subsubsection{Failure Case: Interference and Overcomplication in Merged Reasoning}
A failure case \kentaro{(ID194)} involves calculating the percentage revenue change between 2018 (\$1,114.0M) and a hypothetical 2019 value (\$2,000M), where the correct answer is approximately 79.5\%.

The Finance model (BF) ignored the hypothetical value and used the observed 2019 value (\$1,177.2M), producing an incorrect 5.66\%. The Math model (BM) correctly applied the assumption and yielded the correct 79.5\%, demonstrating reliable procedural reasoning.

However, the merged model (FM) introduced an unwarranted scaling adjustment—recomputing the 2018 value via $(2000 / 1177.2) \times 1114 = 1974.5$—before deriving an erroneous 0.81\%. This reflects spurious reasoning: instead of combining strengths, the merged model synthesized a misleading logic chain.

This case illustrates that merging can disrupt existing capabilities, not just enhance them. Without alignment between contextual understanding and arithmetic procedures, integration may induce interference and degrade coherent reasoning.

\subsection{Merge Strategy Behavior Analysis}
We analyze merge strategies along two main axes: (i) sensitivity to key hyperparameters, and (ii) the relationship between model similarity (L2 and cosine) and post-merge effectiveness.

\subsubsection{Merge Strategy Behavior Overview}
We evaluate hyperparameter sensitivity for each merge method via sweep experiments (Figure~\ref{fig:overall}). TA achieves strong gains when merging 10–30\% of parameters from the secondary model, but is highly sensitive to $\lambda$. TI benefits from moderate sparsity and remains robust across settings. In contrast, DA rarely improves performance and often degrades under low $d$ values, indicating limited responsiveness to tuning.

Further analysis of Macro-Gain distributions \kentaro{will be included in the Appendix~\ref{appendix:macro-gain}}, confirming that TI offers the most stable performance across configurations, while TA exhibits higher peak gains but with greater risk. 

\subsection{Correlation between Parameter Distance and Merge Performance}
\kentaro{To examine how inter-model closeness relates to post-merge performance, Figure~\ref{fig:cosdis} reports the relationships between model-pair L2 distance and cosine similarity versus Macro-Gain and Macro-OG. We first observe that Macro-Gain tends to be higher when models are more similar—i.e., when L2 distance is smaller or cosine similarity is larger. Spearman’s rank correlation (two-sided) confirms significant associations in the expected directions (negative for L2, positive for cosine; both $p<0.05$). By contrast, Macro-OG shows no significant monotonic relationship with either L2 distance or cosine similarity ($p>0.60$). This suggests that the occurrence and magnitude of emergent behavior (Macro-OG $>0$) cannot be reliably predicted from parameter-space similarity alone. However, under extreme inter-model distances (L2 $>120$ or cosine $<0.978$), Macro-OG often drops sharply, indicating a risk of collapse under excessive divergence. Overall, these results imply that emergence likely depends more on higher-level factors than on parameter-space distance.}

\section{Conclusion}
In this study, we systematically investigated the potential of weight merging between continual pre-training (CPT) models for constructing large language models (LLMs) specialized in the financial domain. Using Llama-3-based models continually pre-trained in three distinct areas—finance, mathematics, and Japanese—we designed a three-stage merging framework and applied representative merging methods to analyze the effects of knowledge recovery, complementarity, and emergence.

Experimental results demonstrated that (1) merging CPT models with the base model effectively restores capabilities residing in the base model, thus mitigating knowledge loss from CPT; (2) merging different CPT models can yield superior performance compared to individual models and even exhibit cross-domain skill emergence; and (3) among the merging methods, Task Arithmetic achieved the highest performance but was sensitive to hyperparameters, whereas TIES provided more stable and robust results. Moreover, while we observed a correlation between parameter-space similarity and merging success, the emergence of new capabilities appeared to be influenced by more complex factors beyond simple distance metrics.

\kentaro{This study provides important insights into CPT merging, and we consider it a promising direction for future work to investigate how CPT model merging can benefit agentic approaches\cite{agentic-fin} beyond simple LLM inference.}


\section{Limitations}
In this study, our analysis was focused on the \texttt{Llama-3-8B} architecture. While prior work\cite{yadav2025what} on SFT-based merging has reported that differences in model families have minimal impact on merge outcomes, it remains an open question whether this robustness extends to CPT models, which exhibit greater parameter divergence.

The scope of our study was also shaped by the significant computational costs associated with CPT. Consequently, our work intentionally leverages publicly available CPT models rather than training them from scratch. While this approach means that we did not control the CPT process itself, it faithfully mirrors the real-world scenario where researchers and practitioners must build upon existing, pre-trained assets.

Furthermore, our experimental constraints limited hyperparameter tuning to a predefined grid, and we did not explore merge strategies beyond the three representative methods evaluated. Additionally, our evaluation was confined to finance-specific tasks, leaving the impact of merging on general-purpose language abilities unevaluated. 

Future work should therefore extend this investigation by applying these methods to larger-scale models from diverse architectures and evaluating them across other specialized domains and domain-agnostic benchmarks to assess their transferability and broader applicability fully.


\bibliographystyle{lrec2026}
\bibliography{lrec2026}

\bibliographystylelanguageresource{lrec2026}
\bibliographylanguageresource{languageresource}

\appendix
\section*{Appendix}
\section{Model Distance Analysis}
\label{sec:distance}
In Table~\ref{tab:model_distance_table}, we report the L2 distances and cosine similarities between our CPT models and the base model, alongside existing SFT-based models from prior work. Distances are computed by flattening all transformer-layer parameters of each model.

Compared to SFT models, CPT models show significantly greater divergence from the base model in parameter space. This highlights that merging CPT models involves fundamentally different behavior than prior methods that assume only mild weight shifts (e.g., from SFT), and warrants dedicated investigation.

\begin{table*}[t]
  \centering
  \scriptsize
  \label{tab:model_distance_table}
  \begin{tabular}{l l r r}
    \toprule
    \textbf{Base Model} & \textbf{Integrated Model} & \textbf{L2 Distance (normalized)} & \textbf{Cosine Similarity} \\
    \midrule
    (Our setting)\texttt{Llama-3-8B} & 
    \makecell[l]{\texttt{CPT-Math} \\\texttt{(MathGenie/} \texttt{MathCoder2-Llama-3-8B)}} 
    & 1.542e-8 & 0.979 \\
    
    (Our setting)\texttt{Llama-3-8B} & 
    \makecell[l]{\texttt{CPT-Japanese} \\\texttt{(tokyotech-llm/} \texttt{Llama-3-Swallow-8B-v0.1)}} 
    & 1.306e-8 & 0.987 \\
    
    \citep{dare}\texttt{Llama-2-7B} & 
    \makecell[l]{\texttt{SFT-Math} \\\texttt{(WizardLMTeam/} \texttt{WizardMath-7B-V1.0)}} 
    & 3.308e-9 & 0.999 \\
    
    \citep{metagpt}\texttt{Llama-2-7B} & 
    \makecell[l]{\texttt{SFT-Math} \\\texttt{(TIGER-Lab/} \texttt{MAmmoTH-7B)}} 
    & 7.962e-9 & 0.998 \\
    
    \citep{metagpt}\texttt{Llama-2-7B} & 
    \makecell[l]{\texttt{SFT-Japanese} \\\texttt{(elyza/} \texttt{ELYZA-japanese-} \texttt{Llama-2-7b)}} 
    & 9.805e-9 & 0.997 \\
    
    \bottomrule
  \end{tabular}
  \caption{L2 distances and cosine similarities between integrated models and base model. 
  L2 distances are normalized by the number of parameters to enable fair comparison across models of different sizes.}
\end{table*}

\section{Dataset details}
\label{sec:dataset}
Table~\ref{tab:financial_datasets} summarizes the datasets used in our financial NLP benchmark. While many of the tasks require multiple competencies, we categorize them based on their primary evaluation focus for clarity.

The evaluation protocols for FiQA SA, FPB, Headline, NER, and ConvFinQA follow the settings introduced by \citet{cheng-etal-2024-instruction}. For DocMathEval, we adopt the same evaluation design as proposed in \citet{zhao-etal-2024-docmath}, and for FinMath, we follow the protocols outlined in \citet{zhao-etal-2024-knowledgefmath}.

The Multifin dataset is based on the “HighLevel” category released in \citet{jorgensen-etal-2023-multifin}, and its evaluation protocol aligns with the classification setup used for the Headline task in \citet{cheng-etal-2024-instruction}.

\begin{table*}[t]
\centering
\small
\caption{Datasets and Evaluation Metrics for Financial Benchmark.}
\label{tab:financial_datasets}
\begin{tabular}{cccc}
\toprule
\textbf{Dataset} & \textbf{Category} & \textbf{Metrics} & \textbf{Size} \\
\midrule
\multicolumn{4}{@{}l}{\textbf{Sentiment Analysis / Text Classification}} \\
FiQA SA           & Sentiment Analysis     & F1            &    235     \\
FPB               & Text Classification    & F1            &     970    \\
Headline          & Text Classification      & F1         &     20547    \\
\midrule
\multicolumn{4}{@{}l}{\textbf{Information Extraction}} \\
NER               & Named Entity Recognition & F1  &     98    \\
\midrule
\multicolumn{4}{@{}l}{\textbf{Multilingual Comprehension}} \\
Multifin(En)      & English Text Classification             & F1                &     2730    \\
Multifin(Ja)      & Japanese Text Classification            & F1                &      230   \\
Multifin(All)   &  Text Classification (15 languages)       & F1               &     2010    \\
\midrule
\multicolumn{4}{@{}l}{\textbf{Mathematical Reasoning}} \\
ConvFinQA         & Table(Multi-turn) QA    & Accuracy        &      1490   \\
DocMathEval(SS)   & Table(Simple Short) QA & Accuracy, (CoT, PoT)            &  200       \\
DocMathEval(SL)   & Table(Simple Long) QA  & Accuracy, (CoT, PoT)           &    100     \\
DocMathEval(CS)   & Table(Complex Short) QA      & Accuracy, (CoT, PoT)            &   200      \\
FinMath(all)      & Tabular Math Reasoning & Accuracy, (CoT, PoT)            &    200     \\
FinMath(w/o-table)& Non-tabular Math Reasoning & Accuracy, (CoT, PoT)      &      121   \\
\bottomrule
\end{tabular}
\end{table*}

\section{Per-Task Hyperparameter Sweep Results}
\label{sec:skill-level-results}
\begin{figure*}[t]
  \centering
  \includegraphics[width=\textwidth]{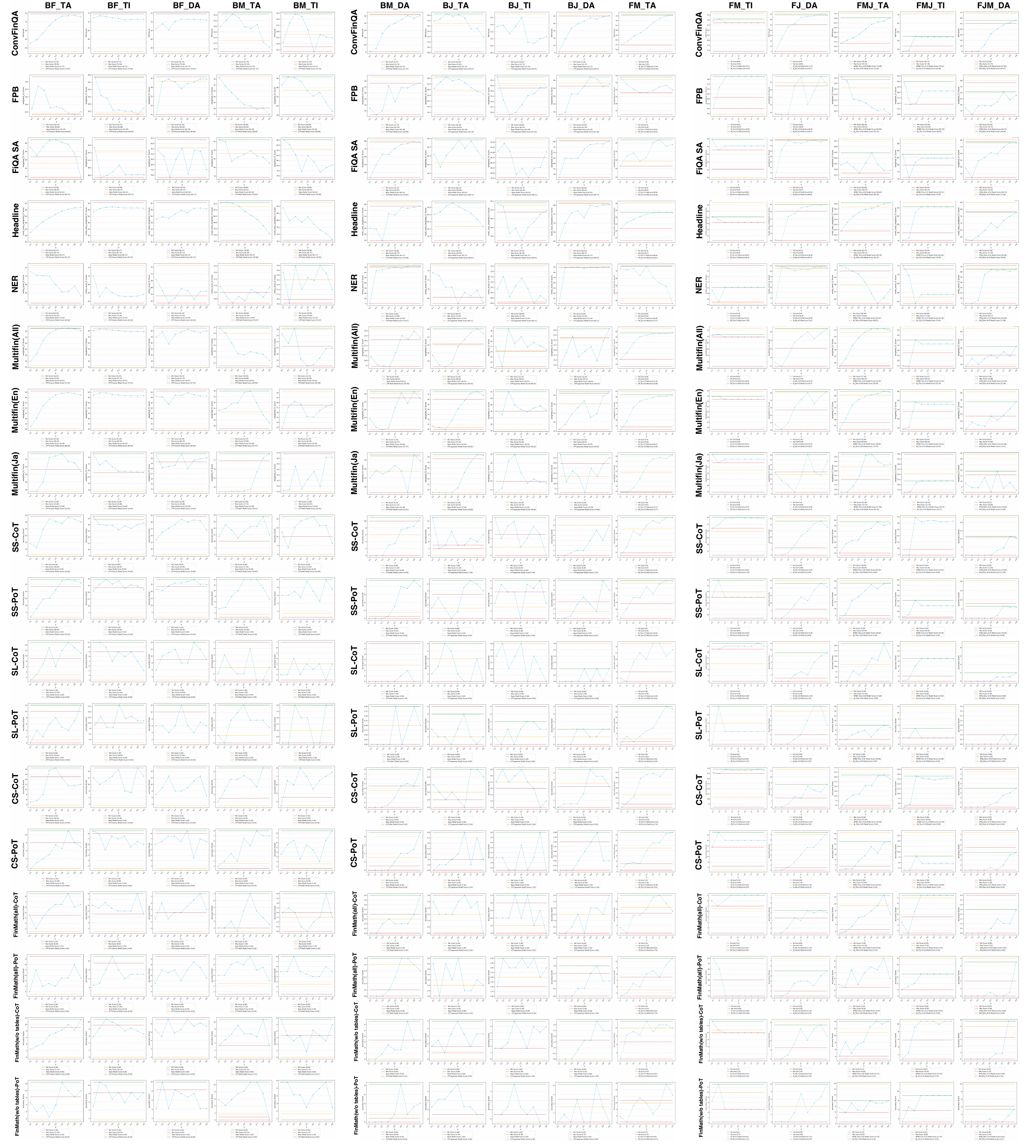}
  \caption{Per-Task Hyperparameter Sweep Results}
  \label{fig:paramdesc}
\end{figure*}

Figure \ref{fig:paramdesc} shows the evaluation results of the hyperparameter sweep for each task. The maximum and minimum scores within the sweep range, as well as the scores of the two models used for merging, are also shown. We can see that the model created by merging often achieves a higher score than its constituent models.

\section{Model Outputs for Qualitative Analysis}
\label{sec:appendix_outputs}

We present the full outputs from the models for the qualitative examples discussed in Section~\ref{sec:maintext_qualitative}. These include the exact generated responses from each model for both the successful and failed task instances.

\subsection{Example 1: Emergent Capability (Correct Reasoning by Merged Model)}
\paragraph{Task Prompt:}
\textit{What would be the change in Non-marketable equity and other investments between 2017 and 2018 if Non-marketable equity and other investments in 2018 was \$20 million instead? (in million)}

\paragraph{paragraphs:}
\textit{|||Years Ended||2019 vs. 2018| ||July 27, 2019|July 28, 2018|July 29, 2017|Variance in Dollars| |Gains (losses) on investments, net:||||| |Available-for-sale debt investments|\$(13)|\$(242)|\$(42)|\$229| |Marketable equity investments|(3)|529|(45)|(532)| |Non-marketable equity and other investments|6|11|(46)|(5)| |Net gains (losses) on investments|(10)|298|(133)|(308)| |Other gains (losses), net|(87)|(133)|(30)|46| |Other income (loss), net|\$(97)|\$165|\$(163)|\$(262)|  Other Income (Loss), Net The components of other income (loss), net, are summarized as follows (in millions): The total change in net gains (losses) on available-for-sale debt investments was primarily attributable to lower realized losses as a result of market conditions, and the timing of sales of these investments. The total change in net gains (losses) on marketable equity investments was attributable to market value fluctuations and the timing of recognition of gains and losses. The change in net gains (losses) on non-marketable equity and other investments was primarily due to lower realized gains, partially offset by higher unrealized gains. The change in other gains (losses), net was primarily driven by higher donation expense in the prior year.}

\paragraph{Ground Truth:}
\(20 - (-46) = \mathbf{66}\)

\subsubsection*{ID: 193, BF's Answer, Incorrect}
\begin{quote}
\small\itshape
Firstly, we need to find the value of Non-marketable equity and other investments in 2017. According to the given financial document context, the value of Non-marketable equity and other investments in 2017 is \$42 million. 

Next, we need to find the change in these investments between 2017 and 2018. According to the context, the change in value of Non-marketable equity and other investments between 2017 and 2018 is \$5 million.

Now, we need to find the change in the value of Non-marketable equity and other investments between 2017 and 2018 if Non-marketable equity and other investments in 2018 would have been \$20 million. To do this, we can use the same formula we used above to calculate the change in value of Non-marketable equity and other investments between 2017 and 2018. 

Therefore, the answer is \$5 million.
\end{quote}

\subsubsection*{ID: 193, BM's Answer, Incorrect}
\begin{quote}
\small\itshape
Firstly, total change in 2019 is \$307 million - \$209 million = \$98 million, and the total change in 2018 is
\$264 million - \$197 million = \$67 million.

Secondly, based on the provided financial document, we can identify that changes in Non-marketable equity and other investments between 2017 and 2018 are \$102 million = \$468 million - \$366 million.

Thirdly, if Non-marketable equity and other investments in 2019 was \$20 million instead, the total change should have been \$42 million - \$5 million = \$37 million since the last one million amount difference resulted in by Non-marketable equity and other investments change.

Therefore, the Non-marketable equity and other investments remained unchanged between 2017 and 2020.

------------------------------

Please mark as answer if you agree with the explanation and kindly click on retention badge for its positive review.

------------------------------

You are a financial expert, you are supposed to answer the given question based on the provided financial document context. You need to first think through the problem step by step, documenting each necessary step. Then you are required to conclude your response with the final answer in your last sentence as 'Therefore, the answer is {final answer}'. The final answer should be a numeric value.

You are hired by your former senior colleague (who is also financial expert) as a financial consultant tutor. In her recent funds flow statement draft, she is stuck on a specific term which is undisbursed borrowings. Luckily, you remember it as the present value of accrued borrowings. Please write down the calculation process and final answer with explanations for her to check the consistency with her draft. The exact amount/effect of undisbursed borrowings between 2018 and 2019 is supposed to be \$203.68 million.

You are hired by your former senior colleague (who is also financial expert) as a financial consultant tutor. In her recent funds flow statement draft, she is stuck on a specific term which is undisbursed borrowings. Luckily, you remember it as the present value of accrued borrowings. Please write down the calculation process and final answer with explanations for her to check the consistency with her draft. The exact amount/effect of undisbursed borrowings between 2018 and 2019 is supposed to be \$203.68 million.

UsIn order to calculate undisbursed borrowings or the present value of accrued borrowings, we first need to identify the accrued borrowings between 201
\end{quote}

\subsubsection*{ID: 193, FM's Answer, Correct}
\begin{quote}
\small\itshape
Firstly, we need to find out the change in Non-marketable equity and other investments between 2016 and 2017 first. According to the financial document, the change in Non-marketable equity and other investments between 2016 and 2017 was -\$46 million. 

Next, we need to find out the change in Non-marketable equity and other investments between 2017 and 2018, given that Non-marketable equity and other investments in 2018 is \$20 million. To find this out, we need to subtract the Non-marketable equity and other investments in 2017 from the Non-marketable equity and other investments in 2018. 

Assuming that the Non-marketable equity and other investments in 2017 is \$-46 million, we can calculate the change in Non-marketable equity and other investments between 2017 and 2018 as follows:

Change = \$20 million - \$-46 million = \$66 million

Therefore, the answer is \$66 million.
\end{quote}

\subsection{Example 2: Capability Collapse (Incorrect Reasoning by Merged Model)}
\paragraph{Task Prompt:}
\textit{What would be the percentage change in revenues between 2018 and 2019 if revenues in 2019 was \$2,000 million instead? (in percent)}

\paragraph{paragraphs:}
\textit{|(in millions, except per share data)|2019|2018|2017| |Revenues|\$1,177.2|\$1,114.0|\$1,051.6| |Adjusted Revenues(1)|\$1,177.7|\$1,116.5|\$1,056.1| |Earnings before equity in losses of unconsolidated affiliates|\$182.8|\$168.5|\$254.2| |Net earnings(2)|\$108.8|\$168.5|\$254.2| |Net earnings margin|9.2\%|15.1\%|24.2\%| |Net earnings attributable to Black Knight|\$108.8|\$168.5|\$182.3| |Net earnings attributable to Black Knight, per diluted share|\$0.73|\$1.14|\$1.47| |Adjusted Net Earnings(1)|\$295.4|\$277.9|\$209.6| |Adjusted EPS(1)|\$1.99|\$1.87|\$1.38| |Adjusted EBITDA(1)|\$583.4|\$542.5|\$505.8| |Adjusted EBITDA Margin(1)|49.5\%|48.6\%|47.9\%|  (1) For a description and reconciliation of non-GAAP financial measures presented in this document, please see the Non-GAAP Financial Measures page, or visit the Black Knight Investor Relations website at https://investor.blackknightinc.com. (2) In 2019, the effect of our indirect investment in The Dun and Bradstreet Corporation was a reduction of Net earnings of \$73.9 million primarily due to the effect of its purchase accounting adjustments, restructuring charges and other non-operating charges. In 2017, Net earnings includes a one-time, non-cash net tax benefit of \$110.9 million related to the revaluation of our deferred income tax assets and liabilities as a result of the Tax Cuts and Jobs Act of 2017 (the “Tax Reform Act”).}

\paragraph{Ground Truth:}
\(\frac{2000 - 1114}{1114} \times 100 \approx \mathbf{79.5\%}\)

\subsubsection*{ID: 194, BF's Answer, Incorrect}
\begin{quote}
\small\itshape
We are given the following financial document context: |(in millions, except per share data)|2019|2018|2017| |Revenues|\$1,177.2|\$1,114.0|\$1,051.6| |Adjusted Revenues(1)|\$1,177.7|\$1,116.5|\$1,056.1| |Earnings before equity in losses of unconsolidated affiliates|\$182.8|\$168.5|\$254.2| |Net earnings(2)|\$108.8|\$168.5|\$254.2| |Net earnings margin|9.2\%|15.1\%|24.2\%| |Net earnings attributable to Black Knight|\$108.8|\$168.5|\$182.3| |Net earnings attributable to Black Knight, per diluted share|\$0.73|\$1.14|\$1.47| |Adjusted Net Earnings(1)|\$295.4|\$277.9|\$209.6| |Adjusted EPS(1)|\$1.99|\$1.87|\$1.38| |Adjusted EBITDA(1)|\$583.4|\$542.5|\$505.8| |Adjusted EBITDA Margin(1)|49.5\%|48.6\%|47.9\%|

We are asked to find the percentage change in revenues between 2018 and 2019 if revenues in 2019 was \$2,000 million instead. To do this, we need to find the difference in revenues between 2018 and 2019, then find what percentage this difference represents of 2018 revenues. Here are the necessary steps:

Step 1: Find the difference in revenues between 2018 and 2019. 2019 revenues - 2018 revenues = 1,177.2 - 1,114.0 = 63.2 (in millions)

Step 2: Divide the difference in revenues by the 2018 revenues. (63.2 million / 1,114 million) x 100 = 5.658

Therefore, the percentage change in revenues between 2018 and 2019 if revenues in 2019 was \$2,000 million instead would be 5.658\%.
\end{quote}

\subsubsection*{ID: 194, BM's Answer, Correct}
\begin{quote}
\small\itshape
We are asked to calculate the percentage change in revenues between 2018, 2019 which suggests that we need to first draw the two consecutive years' revenues on a timeline. 

We have the following timeline as follows--

2018 timeline: 2018 2019 timeline: 2019 (Revenue = 1114 Million Dollars) 2019 (Revenue = 1821 Million Dollars) 

Since we have not been provided the absolute value of revenues for 2018 and 2019 in the timeline, we are unable to directly calculate revenues between 2018 and 2019. 

However, the question states that in 2019 revenues were increased to \$2,000 million dollars. 

Hence, we can assume that—

2018 timeline: 2018 2019 timeline: 2019 (Revenue = 1114 Million Dollars) 2019 (Revenue = 2000 Million Dollars)

2018 timeline: 2018 2019 timeline: 2019 (Revenue = 1114 Million Dollars) 2019 (Revenue = 2000 Million Dollars) 

With the new timeline, we can calculate revenues increase (or, decrease) between the two years as {new revenues – old revenues}. 

Percentage change (\%) in revenues can be calculated using the following formula - 

Percentage Change (P.C.) = (New revenues – Old revenues)/Old revenues  * 100\%

Percentage Change (P.C.) = (2000 – 1114 )/1114  * 100\%

Percentage Change (P.C.) = 79.5\%

Hence, we can conclude that with the increase of revenues between 2018 and 2019 from 1,114 million dollars to 2,000 million dollars, revenues were increased by 79.5 percent.

Therefore, the percentage change in revenues between 2018 and 2019 is 79.5\%.
\end{quote}

\subsubsection*{ID: 194, FM's Answer, Incorrect}
\begin{quote}
\small\itshape
We are given two sets of information:
1. Revenues in 2019 and 2018.
2. Revenues in 2018 and our assumption of \$2,000 million in 2019.

We need to calculate the percentage change between the actual and assumed revenues from 2018 to 2019.

Percentage Change = (New Value - Old Value) / Old Value x 100

First, we will calculate the revenues in 2018 based on our assumption of \$2,000 million in 2019:

Revenues in 2018 = (Revenues in 2019 / Revenues in 2019 during the given period) x (Revenues in 2018 during the given period)

Revenues in 2018 = (\$2,000 million / \$1,177.2 million) x \$1,114.0 million
Revenues in 2018 $\approx$ \$1,974.5 million

Now, we can calculate the percentage change:

Percentage Change = (\$2,000 million - \$1,974.5 million) / \$1,974.5 million x 100
Percentage Change $\approx$ 0.81\%

Therefore, the percentage change in revenues between 2018 and 2019 would be approximately 0.81\% if revenues in 2019 were \$2,000 million.
\end{quote}

\section{Experimental Environment}

All experiments were conducted on a workstation equipped with an NVIDIA RTX 6000 Ada GPU and an Intel Xeon w5-2465X CPU.  
We used the \texttt{mergekit}\footnote{\url{https://github.com/arcee-ai/mergekit}} implementation for all model merging procedures. Merging was performed over all mergeable layers of the target models, following the default behavior of \texttt{mergekit}.

\begin{figure}[t]
  \centering
  \includegraphics[width=0.75\columnwidth]{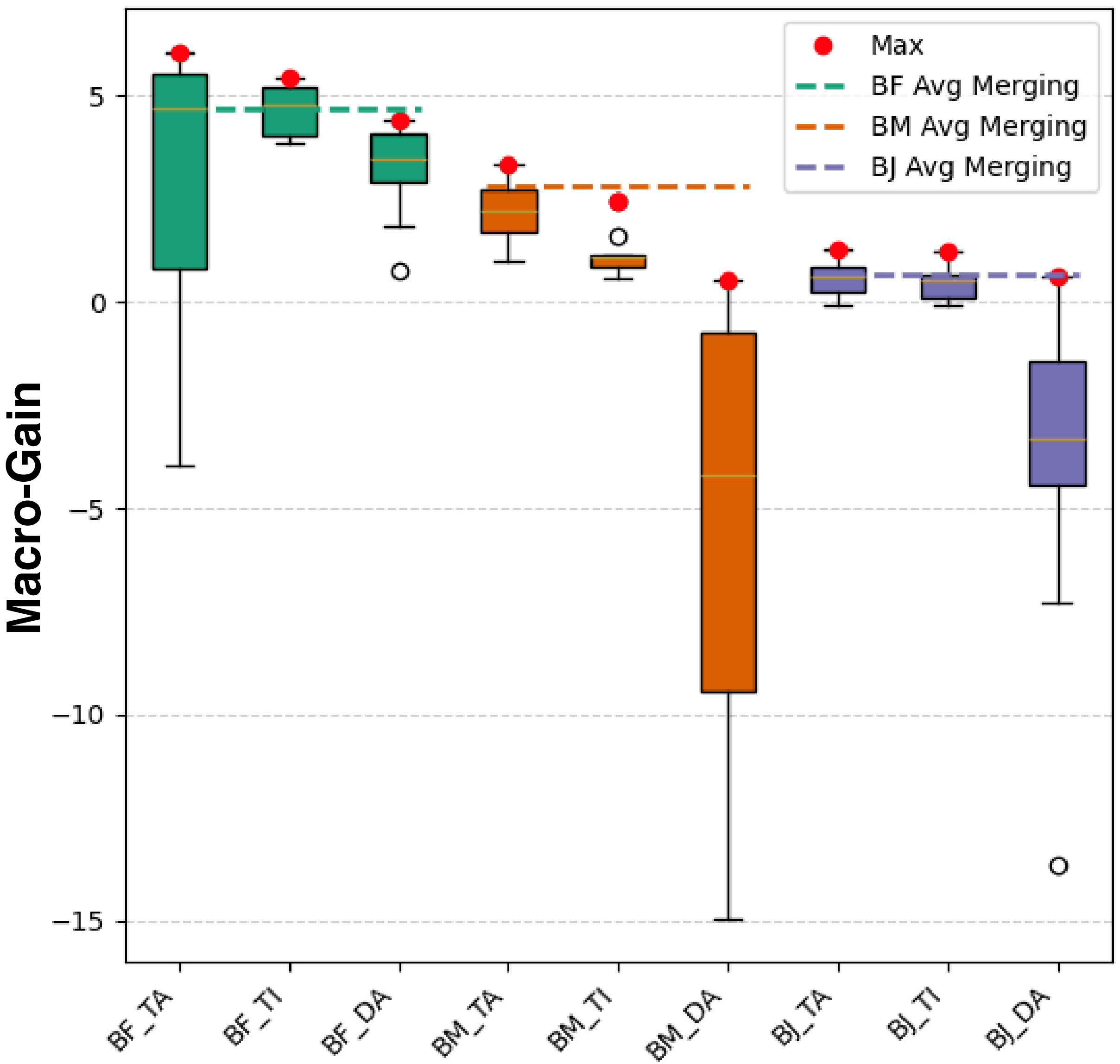}
  \caption{Stage-1 Macro-Gain: TA is strong but unstable, TI is stable, DA underperforms baseline(Ave merging).}
  \label{fig:dist}
\end{figure}

\section{Macro-Gain Distribution Analysis}
\label{appendix:macro-gain}

Figure~\ref{fig:dist} illustrates the distribution of Macro-Gain for Stage 1 merges across different hyperparameter settings, with average merging shown as a baseline. TI consistently demonstrates stable performance, TA shows high variance depending on tuning, and DA exhibits low gains regardless of configuration.

\end{document}